%% file: sample-sigconf.tex
\documentclass[sigconf]{acmart}
\AtBeginDocument{%
  }

\usepackage{balance}
\usepackage{bm}
\usepackage{multirow}
\usepackage{amsmath}
\usepackage{amsthm}
\usepackage{enumerate}
\usepackage{enumitem}
\usepackage{subfigure}
\usepackage{stfloats}
\usepackage{hyperref}
\usepackage{bbding}
\usepackage{algorithmicx}
\usepackage{algorithm}
\usepackage{appendix}

\newtheorem{remark}{Remark}
\newtheorem{theorem}{Theorem}
\newtheorem{lemma}{Lemma}
\usepackage{graphicx}

\begin{document}

\title{Privacy-Preserving Orthogonal Aggregation for Guaranteeing Gender Fairness in Federated Recommendation}

\author{Siqing Zhang}
\orcid{0009-0009-0518-5888}
\affiliation{%
  \institution{University of Science and Technology of China}
  \department{CCCD Key Lab of Ministry of Culture and Tourism}
  \city{Hefei}
  \country{China}
}
\email{siqingzhang@mail.ustc.edu.cn}

\author{Yuchen Ding}
\orcid{0009-0008-3193-1082}
\affiliation{%
  \institution{University of Science and Technology of China}
  \department{CCCD Key Lab of Ministry of Culture and Tourism}
  \city{Hefei}
  \country{China}
}
\email{yuchending@mail.ustc.edu.cn}

\author{Wei Tang}
\orcid{0000-0001-6561-7026}
\affiliation{%
  \institution{University of Science and Technology of China}
  \department{CCCD Key Lab of Ministry of Culture and Tourism}
  \city{Hefei}
  \country{China}
}
\email{weitang@mail.ustc.edu.cn}

\author{Wei Sun}
\orcid{0000-0002-7319-004X}
\affiliation{%
  \institution{University of Science and Technology of China}
  \department{CCCD Key Lab of Ministry of Culture and Tourism}
  \city{Hefei}
  \country{China}
}
\email{sunw@ustc.edu.cn}

\author{Yong Liao}
\authornote{co-corresponding authors}
\orcid{0000-0001-6403-0557}
\affiliation{%
  \institution{University of Science and Technology of China}
  \department{CCCD Key Lab of Ministry of Culture and Tourism}
  \city{Hefei}
  \country{China}
}
\email{yliao@ustc.edu.cn}

\author{Peng Yuan Zhou}
\authornotemark[1]
\orcid{0000-0002-7909-4059}
\affiliation{%
  \institution{Aarhus University}
  \department{Department of Electrical and Computer Engineering}
  \city{Aarhus}
  \country{Demark}
}
\email{pengyuan.zhou@ece.au.dk}
\renewcommand{\shortauthors}{Siqing Zhang et al.}

\begin{abstract}
Under stringent privacy constraints, whether federated recommendation systems can achieve group fairness remains an inadequately explored question. Taking gender fairness as a representative issue, we identify three phenomena in federated recommendation systems: performance difference, data imbalance, and preference disparity. We discover that the state-of-the-art methods only focus on the first phenomenon. Consequently, their imposition of inappropriate fairness constraints detrimentally affects the model training. Moreover, due to insufficient sensitive attribute protection of existing works, we can infer the gender of all users with 99.90\% accuracy even with the addition of maximal noise. In this work, we propose \textbf{P}rivacy-\textbf{P}reserving \textbf{O}rthogonal \textbf{A}ggregation (\textbf{PPOA}), which employs the secure aggregation scheme and quantization technique, to \textbf{prevent the suppression of minority groups by the majority and preserve the distinct preferences} for better group fairness. PPOA can assist different groups in obtaining their respective model aggregation results through a designed orthogonal mapping while keeping their attributes private. Experimental results on three real-world datasets demonstrate that PPOA enhances recommendation effectiveness for both females and males by up to 8.25\% and 6.36\%, respectively, with a maximum overall improvement of 7.30\%, and achieves optimal fairness in most cases. Extensive ablation experiments and visualizations indicate that PPOA successfully maintains preferences for different gender groups.
\end{abstract}



\maketitle

\input{sections/1_intro.tex}
\input{sections/2_relatedworks.tex}
\input{sections/3_preliminaries.tex}

\input{sections/4_analysis.tex}

\input{sections/5_method.tex}
\input{sections/6_evalution.tex}
\input{sections/7_discussion.tex}

\begin{acks}
This work is supported by the National Key Research and Development Program of China (2022YFB3105405, 2021YFC3300502) and the Provincial Key Research and Development Program of Anhui (202423110050033)
\end{acks}

\newpage
\clearpage
\bibliographystyle{ACM-Reference-Format}
\balance
\bibliography{bibfile}

\input{sections/ethicalconsiderations}
\input{sections/appendix}
\end{document}

%% file: sections/1_intro.tex
\section{\MakeUppercase{Introduction}} 
Traditional recommendation models necessitate the aggregation of user data, posing risks to user privacy and potentially breaching legal regulations (e.g. GDPR~\cite{voigt2017eu}). As a paradigm of privacy-preserving distributed machine learning, \textit{Federated Learning}~\cite{mcmahan2017communication} (FL) leads to the emerging significance of \textit{Federated Recommendation Systems} (FRSs)~\cite{chai2020secure,perifanis2022federated,zhang2023dual,li2023federated}. In FRSs, users train local models and upload item embeddings to a central server. The server then aggregates these updates and distributes the new global model.

\begin{figure*}[t!]
    \centering
    \includegraphics[width=\textwidth]{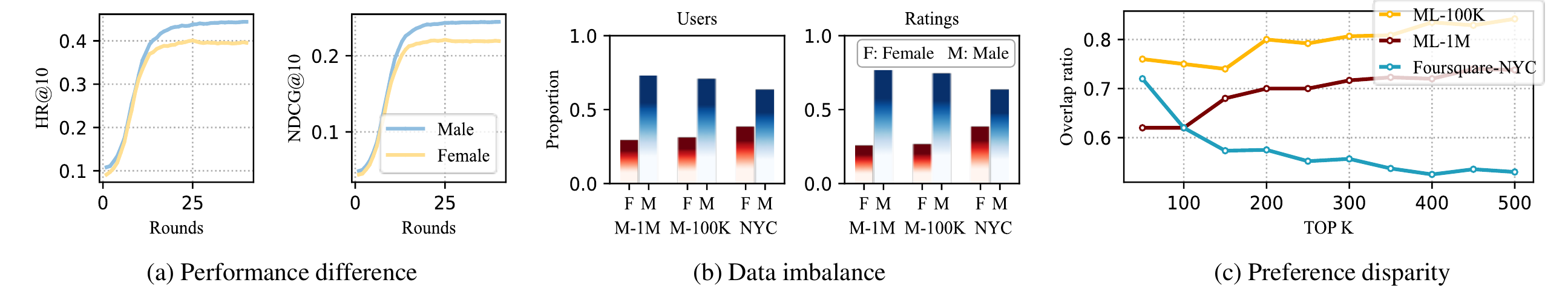}
    \caption{Three gender-related phenomena in FRS.}
    \label{fig:ex0-genderbias}
\end{figure*}

The distributed machine learning paradigm presents challenges in achieving group fairness within recommendation systems.
Group fairness is described as the principle that models should not ignore the preferences of minority groups~\cite{malecek2021fairness}, and minority groups should not be biased by the training data of the majority groups~\cite{fu2020fairness}. 
In prior works, fairness-aware strategies are often designed and implemented within centralized settings~\cite{LiGZ21,wang2023survey}. However, as many users are reluctant to disclose their sensitive attributes, such as gender and other demographic information, these methods face limitations when applied directly to federated settings where the importance of user privacy concerns is heightened. In this work, we focus on gender fairness, a widely discussed and particularly representative problem~\cite{ekstrand2012fairness,geyik2019fairness,jin2023survey}, as a critical entry point for examining group fairness. We identify three phenomena that potentially lead to gender unfairness: \textbf{(1)} \textit{Performance difference}: as depicted in Fig.~\ref{fig:ex0-genderbias}(a), FedMF~\cite{chai2020secure}, a fundamental FRS, yields better recommendations for male users compared to female users on ML-1M dataset, \textbf{(2)} \textit{Data imbalance}: as shown in Fig.~\ref{fig:ex0-genderbias}(b), there are significant differences in the number of users and interaction records for each gender in the datasets, and \textbf{(3)} \textit{Preference disparity}: the disparity in preferences for the top K items favored by each gender reaches up to nearly 50\%, as illustrated in Fig.\ref{fig:ex0-genderbias}(c).

F2MF~\cite{liu2022fairness} claims to be the first to implement fair-aware federated matrix factorization. Building on F2MF, a recent work F$^2$PGNN~\cite{agrawal2024no} introduces higher-order information to achieve group fairness. 
However, both F2MF and F$^2$PGNN do not consider the data imbalance and preference disparity between groups, seeking fairness merely by aligning the learning speeds of the advantaged group with the disadvantaged group\footnote{``\textit{The low-performance group needs to learn faster and the high-performance group needs to learn slower in order to produce a better group level fairness.}''~\cite{liu2022fairness}} to eliminate performance differences. Consequently, these straightforward constraints suppress the performance of the advantaged group (as shown in Fig.~\ref{fig:maincomp}), which often contribute more to the dataset, and the aggregation method could lead to minority groups receiving model updates that are overly influenced by the preferences of the advantaged groups, thereby biasing the training process.
Moreover, we find existing noise-based methods used in F2MF and F$^2$PGNN pose a risk of privacy leakage in protecting sensitive attributes. Specifically, we devise an attack method that can determine the gender of all users with 99.90\% accuracy even with the addition of maximal noise (as described in Sec.~\ref{sec:anasis}). 

In this work, we explore a different approach to achieve better group fairness. Firstly, we identify the federated aggregation phase as the critical juncture, where preferences of the minority are often submerged by those of the dominant, e.g., the smaller number of female users (data imbalance) results in male-related information predominating in the aggregated parameters. Hence, preventing the inter-group interference during the training phase can improve group fairness, especially for the minority.   
Based on these observations, we propose \textbf{O}rthogonal \textbf{A}ggregation (\textbf{OA}), which helps users obtain aggregation results for their respective groups, thus preserving the distinct preferences and preventing the mutual suppression.
Furthermore, to counteract the vulnerabilities of existing privacy protection methods, we introduce a \textit{Secure Aggregation} (SA) scheme, by which users mask their model updates and utilize \textit{Quantization}~\cite{zheng2022aggregation} to reduce communication overhead. When directly applying SA, the model updates received by the server is indistinguishable from random vectors, making it impossible to enforce existing fairness constraints. However, OA can seamlessly integrate with SA, i.e., \textbf{P}rivacy-\textbf{P}reserving \textbf{O}rthogonal \textbf{A}ggregation (\textbf{PPOA}), which not only ensures privacy but also enables fair recommendation.

Our contributions can be summarized as follows:
\begin{itemize}[leftmargin=*]
    \item We uncover the insufficiency of existing methods in terms of protecting the privacy of sensitive attributes. Despite the addition of the maximum noise derived from F2MF, we can infer the gender of all users with 99.90\% accuracy.
    \item We summarize 3 phenomena that can lead to gender unfairness and propose PPOA to preserve preferences of all groups, filling the gap in protecting sensitive attribute privacy when achieving group fairness of FRSs.
    \item Detailed theoretical analysis proves the correctness of PPOA. Extensive experiments conducted on three real-world datasets demonstrate the superiority of PPOA.
\end{itemize}

%% file: sections/2_relatedworks.tex
\section{\MakeUppercase{related works}}
\label{sec:relatedworks}
For FRSs, several investigations~\cite{LinLPM21,LiangP021,luo2022personalized} dedicate efforts towards explicit feedback that demands stringent requirements on user interaction data. Other studies~\cite{flanagan2021federated,luo2022personalized,agrawal2024no} access additional data sources during modeling, leveraging higher-order information of users and items. Our work primarily focus on widely applied implicit feedback and foundational scenarios that refrain from utilizing supplementary data sources, aligning with many FRSs~\cite{chai2020secure,perifanis2022federated,zhang2023dual,li2023federated}.

\textbf{Fair FRSs}. Several works~\cite{yu2020sustainable,ZhaoJ22,SuYWLL024} focus on client fairness, aiming to motivate clients to contribute resources and prevent profit inequity. RF$^2$~\cite{maeng2022towards} identifies the interdependence of system and data heterogeneity within FRSs and explores its impact on fairness. 
F2MF~\cite{liu2022fairness} is the first to investigate group fairness in FRS, asserting that achieving such fairness necessitates consistent model performance across different groups. F$^2$PGNN~\cite{agrawal2024no} extends F2MF to scenarios involving multiple data sources. The fairness constraints established by F2MF has been applied ever since. However, simply imposing constraints to equalize performance can undermine the model performance of advantaged groups, which contradicts the principle of Pareto optimization~\cite{censor1977pareto}. We aim to achieve group fairness \textbf{without compromising the interests of any party}.

\textbf{Privacy-Preserving FRSs}. FedMF~\cite{chai2020secure} posits that even when user embeddings are kept locally, servers can infer user interaction records through received item embeddings, leading to privacy leakage. It employs \textit{Homomorphic Encryption} (HE) to preserve privacy~\cite{aono2017privacy}. Meanwhile, the majority of FRSs~\cite{liu2022fairness,zhang2023dual,li2023federated,agrawal2024no} rely on \textit{Differential Privacy} (DP) to protect privacy for better efficiency. However, we highlight the failure of existing methods in preserving sensitive attributes (see details in Sec.~\ref{sec:anasis}). Therefore, in this work, we adopt a mask-based secure aggregation scheme~\cite{Bonawitz2017PracticalSA,Liu2022EfficientDA,zheng2022aggregation}, where users add \textit{One-time Pads} to their original vectors to conceal privacy, and servers automatically offset the masks during the aggregation process. Such approaches pose less privacy leakage threats compared to DP and incur lower overhead than HE. Similar to many prior works~\cite{wu2022federated,Wu2022FPDAFA}, the generation of masks in this work is delegated to a trusted third-party ($\mathcal{TTP}$) to mitigate user overhead, and $\mathcal{TTP}$ does not have access to users' gender information or model updates.

%% file: sections/3_preliminaries.tex
\section{\MakeUppercase{PRELIMINARIES}}
\label{sec:preliminaries}
\subsection{Problem Formulation and Notations}
$\mathbf{R}=[r_{i,j}]\in \{0,1\}^{n\times m}$ denotes the interaction matrix between $n$ users and $m$ items, where $r_{i,j}$ represents the rating given by user $u_i$ to item $j$. For each user $u_i,i\in[1,n]$, $\mathbf{D}_i$ signifies the local data of $u_i$, $\mathbf{U}_i$ and $\mathbf{I}_i$ denotes the user embedding and local item embeddings of $u_i$, respectively. The loss function of $u_i$ is:
\begin{equation}
    \label{eq:rec_loss}
    \mathcal{L}_i \left(\mathbf{U_i}, \mathbf{I_i}\right) =   \sum_{(i, j) \in \mathbf{D_i}}-\left(r_{i,j} \log \hat{r}_{i,j}+\left(1-r_{i,j}\right) \log \left(1-\hat{r}_{i,j}\right)\right),
\end{equation}
where $\hat{r}_{i,j}$ denotes the predicted $r_{i,j}$.
During the $t$-th aggregation, each user uploads $\bm{\Theta}_i^t$, usually containing local item embeddings, to the server. Generally, the server aggregates updates by:
\begin{equation}
    \bm{\Theta}^{t+1} = \sum_{i=1}^n {\omega_i \bm{\Theta}_i^t},\quad s.t.\quad \sum_{i=1}^n {\omega_i=1}.
\end{equation}

The overall objective function in FRS is:
\begin{equation}
    \min_{\{\bm{\Theta_i}\}_{i\in[1,n]}} \sum_{i=1}^n \mathcal{L}_i.
\end{equation}

\subsection{Secure Aggregation and Quantization}
\label{sec:sapQ}
In SecAgg~\cite{Bonawitz2017PracticalSA}, the state-of-the-art secure aggregation scheme, $u_i$ performs the following calculation for the $l$-th entry of the original vector $\bm{\Theta_i}=[\theta_1^i,\theta_2^i,\ldots,\theta_d^i]$:
\begin{equation}
    \vartheta_l^i=\theta_l^i  +\underbrace{\operatorname{PRG}(b_i) + \sum_{o \in \mathcal{U}: i<o} \operatorname{PRG}(s_{i, o}) - \sum_{o \in \mathcal{U}: i>o} \operatorname{PRG}(s_{o, i})}_{\bm{\xi_i}},
\end{equation}
where PRG is a pseudorandom number generator, $b_i$ is a user secret key, $s_{i, o}$ is a secret key negotiated between $o$ and $i$, $\mathcal{U}$ is the entire user set and $\bm{\xi_i}$ is the part generated by $\mathcal{TTP}$ for $i$. SecAgg requires elements to be in a finite field.

The quantization technique~\cite{zheng2022aggregation} helps reduce communication overhead for users in federated systems. For any $p\in [-\kappa ,\kappa]$, $sgn(p)$ denotes the sign of $p$, $round$ denotes the standard rounding of $p$ and $abs(p)$ is the absolute value of $p$. A quantizer $\mathcal{Q}_h$ quantizes $p$ to a h-bit integer in $[-(2^{h-1}-1),2^{h-1}-1]$:
\begin{equation}
    \mathcal{Q}_h(p)=\operatorname{sgn}(p) \cdot \operatorname{round}\left(\operatorname{abs}(p) \cdot\left(2^{h-1}-1\right) / \kappa \right).
\end{equation}

The de-quantizer $\mathcal{Q}_h^{-1}$ can recover a quantized value $q$:
\begin{equation}
    \mathcal{Q}_h^{-1}(q)=\operatorname{sgn}(q) \cdot\left(\operatorname{abs}(q) \cdot \kappa  /\left(2^{h-1}-1\right)\right).
\end{equation}

If $h=16$, after quantization, the communication overhead between users and the server is reduced by half compared to the original transmission of 32-bit floating-point numbers.

%% file: sections/4_analysis.tex
\section{\MakeUppercase{Sensitive attribute inference attack}}
\label{sec:anasis}
In this section, we briefly introduce the method for achieving group fairness of F2MF~\cite{liu2022fairness} and F$^2$PGNN~\cite{agrawal2024no}, the state-of-the-art works, focusing on their vulnerabilities concerning the leakage of user-sensitive attributes. Furthermore, we conduct empirical experiments to validate these findings.

\subsection{Existing Methods for Group Fairness}
The loss function of users in F2MF and F$^2$PGNN is:
\begin{equation}
    \mathcal{L} = \mathcal{L}_{rec} + \lambda \mathcal{L}_{fair},
\end{equation}
where $\mathcal{L}_{rec}$ is similar to Eq.~\eqref{eq:rec_loss}, $\lambda$ is a trade-off parameter and $\mathcal{L}_{fair}$ is:
\begin{equation}
    \label{eq:f2mf_loss}
    \mathcal{L}_{fair}\left(G_0, G_1, \mathcal{E}\right)=\left|\underbrace{\frac{1}{\left|G_0\right|} \sum_{u \in G_0} \mathcal{E}(u)}_{A}-\underbrace{\frac{1}{\left|G_1\right|} \sum_{u \in G_1} \mathcal{E}(u)}_B\right|^\rho,
\end{equation}
where $G_0$ and $G_1$ represent two groups (e.g. male and female), $\mathcal{E}$ is the recommended performance metric function ($\mathcal{E}=1- \mathcal L_{rec}$ in this paper) and $\rho \in \{1,2\}$ determines the smoothness.
From Eq.~\eqref{eq:f2mf_loss}, we can see that the idea of F2MF is to forcibly add constraints to align the performance of two groups. 

When conducting federated aggregation, users need to upload updates as follows:
\begin{equation}
\label{eq:f2mf_grad}
    \begin{aligned}
        \nabla \bm{\Theta_i} & =\frac{\partial}{\partial \bm{\Theta_i}} \mathcal{L}_{ {rec }}^i+\lambda \frac{\partial}{\partial \bm{\Theta_i}} \mathcal{L}_{ {fair }} \\
                             & = \underbrace{\left( 1-\lambda C|A-B|^{\rho-1} \right)}_D\frac{\partial}{\partial \bm{\Theta_i}} \mathcal{L}_{ {rec}}^i,
    \end{aligned}
\end{equation}
where $C=\rho(-1)^{\mathbb{I}(A<B)}(-1)^{\mathbb{I}\left(u \notin G_0\right)}$ and $\mathbb{I}(\cdot)$ denotes the indicator function, with values of $1$ and $0$ respectively when $\cdot$ is true and false. $D$ represents the ratio of learning speed compared to the original.

Therefore, in addition to uploading basic model updates, user $u$ also needs to upload:
\begin{equation}
    \begin{aligned}
         & \nabla A_{ {sum }} | u    =\mathbb{I}\left(u \in G_0\right) \mathcal{E}_u+\epsilon_{1, u}+\epsilon_{A, t}, \\
         & \nabla B_{ {sum }} |u   =\mathbb{I}\left(u \in G_1\right) \mathcal{E}_u+\epsilon_{2, u}+\epsilon_{B, t},   \\
         & \nabla A_{ {count }} | u =\mathbb{I}\left(u \in G_0\right)+\epsilon_{3,u} ,                                \\
         & \nabla B_{ {count }} | u  =\mathbb{I}\left(u \in G_1\right)+\epsilon_{4, u},
    \end{aligned}
\end{equation}
where $\epsilon_{1, u},\epsilon_{A, t},\epsilon_{2, u},\epsilon_{B, t},\epsilon_{3,u},\epsilon_{4, u} \sim \mathcal{N}(0,\sigma)$ are the noise-based privacy protection measures. Below we demonstrate why the noise is insufficient to protect the privacy of sensitive attribute information.

\subsection{Privacy Leakage Analysis}
\label{sec:f2mf_attack}
Taking a user $u \in G_0$ as an example, we focus on the two parameters that need to be uploaded:
\begin{equation}
    \begin{aligned}
         & \nabla A_{ {count }} | u =1+\epsilon_{3,u},  \\
         & \nabla B_{ {count }} | u  =\epsilon_{4, u},
    \end{aligned}
\end{equation}
where $\epsilon_{3,u},\epsilon_{4, u} \sim \mathcal{N}(0,\sigma)$. If $\nabla A_{ {count }} | u$ and $\nabla B_{ {count }} | u$ can be distinguished, then the gender of $u$ can be determined.
\begin{lemma}
    \label{ana:lemma1}
    Given $X$ is an observation from a normally distributed random variable, i.e. $X \sim \mathcal{N}(\mu,\sigma)$, then:
    \begin{equation}
        \operatorname{Pr}(\mu-3 \sigma \leq X \leq \mu+3 \sigma) \approx 99.73 \%.
    \end{equation}
\end{lemma}

\begin{figure}[h!]
    \centering
    \includegraphics[width=0.6\linewidth]{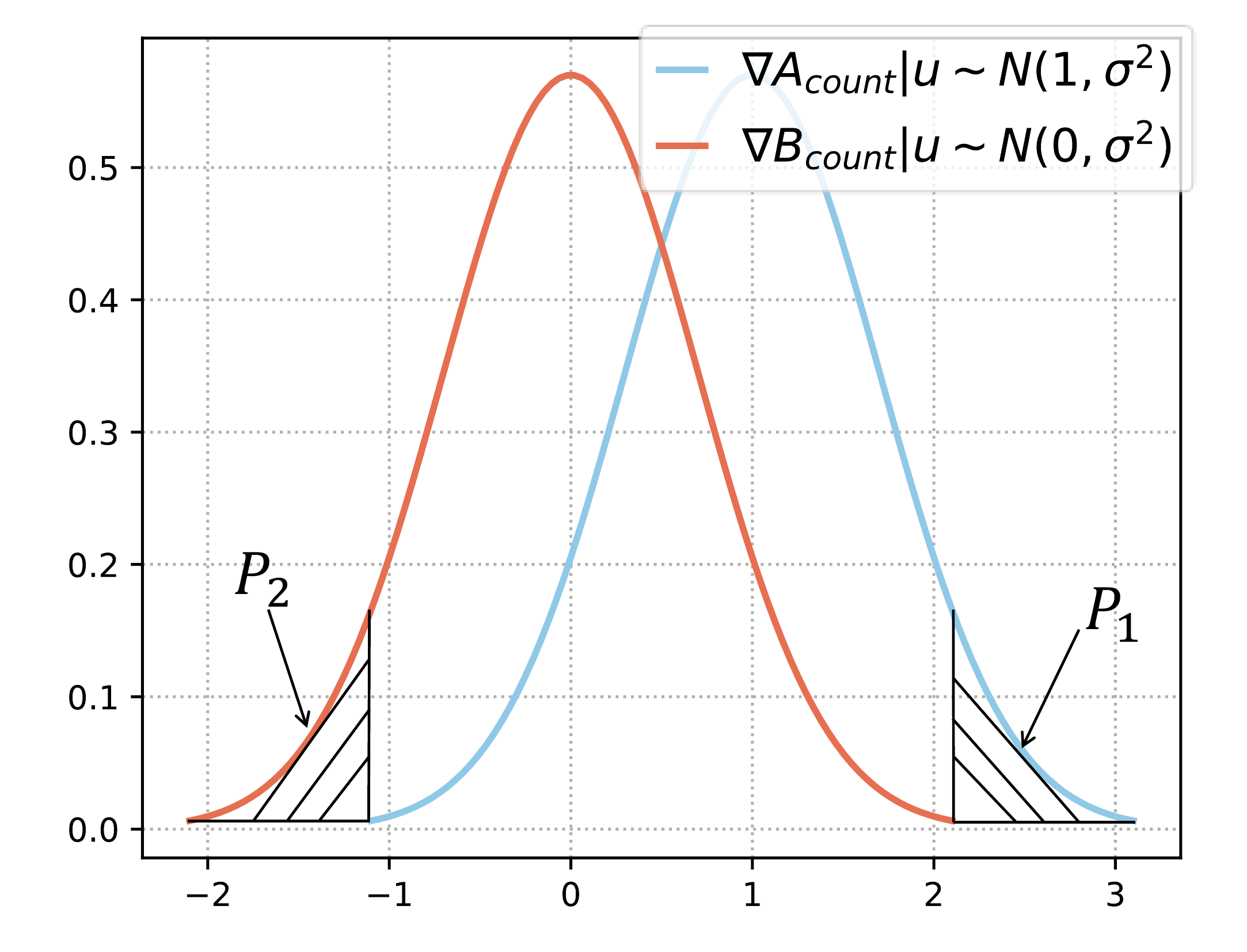}
    \caption{A diagram for identifying uploaded data of exposed users.}
    \label{fig:attack_demo}
\end{figure}

From Lemma~\ref{ana:lemma1} (see Appendix~\ref{app:lemma1} for the proof of it), we deduce that if $\nabla A_{ {count }} | u$ exceeds $3\sigma$, theoretically, there is a $99.865\%(=1-\frac{1-0.9973}{2})$ confidence level that $u \in G_0$. Similarly, if user $u$'s uploaded $\nabla B_{ {count }} | u<1-3\sigma$, we can assert with the same level of confidence that $u \in G_0$. Users whose gender can be determined in the aforementioned strategy is referred to as ``\textit{Exposed Users}''. Let $\Phi(x,\mu,\sigma)$ denote the cumulative distribution function of $\mathcal{N}(\mu,\sigma)$. The uploaded data of exposed users is illustrated in Fig.~\ref{fig:attack_demo}.

In theory, the percentage of exposed users $P_{su}(\sigma)$ is:
\begin{equation}
    P_{su}(\sigma)=1-(1-P_1)(1-P_2),where \left\{\begin{matrix}
         & P_1=1-\Phi(3\sigma,1,\sigma) \\
         & P_2=\Phi(1-3\sigma,0,\sigma)
    \end{matrix}\right..
\end{equation}

\begin{figure}[t!]
    \centering
    \includegraphics[width=0.9\linewidth]{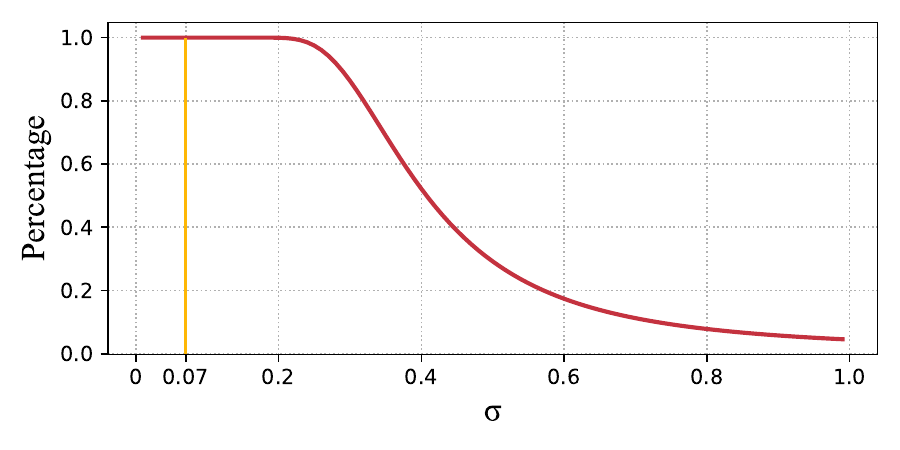}
    \caption{The percentage of exposed users among all users in theory.}
    \label{fig:attack_theo}
\end{figure}

The lower bound and the upper bound of $\sigma$ given in \cite{liu2022fairness} is:
\begin{equation}
    \frac{\mathcal{F}_u}{\sqrt{2} \Phi^{-1}\left(0.5+\delta_1\right)} \leq \sigma \leq H\left|\bar{X}_{\text {actual }}\right| \sqrt{n \delta_2},
\end{equation}
where $H$ is the ratio between the difference and the absolute value of group-wise
performances (see \textbf{A.2} of \cite{liu2022fairness}), $\bar{X}_{\text {actual}}$ is the average ground-truth value, $n$ is the number of users, $\delta_1$ and $\delta_2$ are small constants ($\delta_1=0.1,H=\delta_2=0.01$ in \cite{liu2022fairness}). Take $\bar{X}_{\text {actual}}=1$, then the $\sigma_{max}$ of $\epsilon_{3,u},\epsilon_{4, u}$ is 0.07. 
Fig. \ref{fig:attack_theo} provides the theoretical percentage of exposed users among all users. We believe that \textbf{the lower bound of $\sigma$ given by F2MF is not sufficient to protect the privacy of sensitive attributes}. It can be seen that even when $\sigma=\sigma_{max}=0.07$, almost all users can be identified as exposed users. Next, we conduct practical experiments to confirm it.

\subsection{Attack Experiments}
\textbf{Setup.} Consistent with the setting of F2MF, we use the MovieLens-1M\footnote{https://grouplens.org/datasets/movielens/1m/} (ML-1M) dataset, retain users with more than 20 interactions, resulting in a total of 6,022 training users, and adopt 80\%-10\%-10\% split for each user data based on temporal order. The sensitive attribute considered is gender, with $G_0$ denoting the male group. The value of $\sigma$ is set to $\{0.07,0.1,0.2,0.8,1.0\}$. Upon receiving model updates, the server implement the attack strategy outlined in Sec.~\ref{sec:f2mf_attack} to infer each user's gender, and identify exposed users along with their presumed genders. Then we verify the attack results based on the ground truth and obtain the attack accuracy.

\textbf{Results.} Tab.~\ref{tab:attack_ex} summarizes the percentage of users whose gender could be inferred (i.e. $\frac{|\text{Exposed users}|}{|\text{Total users}|}$), and the accuracy, calculated as $\frac{|\text{Correctly inferred users}|}{|\text{Exposed users}|}$. It is evident that the actual experimental outcomes closely align with the theoretical percentage of exposed users at corresponding $\sigma$ values depicted in Fig.~\ref{fig:attack_theo}. The results demonstrate that even when noise exceeding the upper bound is added in accordance with F2MF, the gender of a substantial number of users can still be accurately inferred, posing a serious privacy leakage issue. 
\input{tab/attack_ex.tex}

%% file: tab/attack_ex.tex
\begin{table}[h!]
    \caption{
        Vulnerabilities with different $\sigma$.}
    \centering
    \resizebox{\linewidth}{!}{
        \begin{tabular}{ccccccc}
            \toprule
            {$\sigma$}           &0      & 0.07             & 0.1              & 0.2              & 0.8              & 1.0              \\
            \midrule[0.5pt]

        {Suspected users}     &6022     & 6022             & 6022             & 6020             & 474              & 286              \\
            Percentage          &100.0\%       & 100.0\%          & 100.0\%          & 99.97\%          & 7.87\%           & 4.75\%           \\
            {Correctly inferred users} &6022& 6016             & 6016             & 6010             & 458              & 270              \\
            Accuracy           & \textbf{100.0\%}        & \textbf{99.90\%} & \textbf{99.90\%} & \textbf{99.83\%} & \textbf{96.62\%} & \textbf{94.41\%} \\
            \bottomrule
        \end{tabular}}
    \label{tab:attack_ex}
\end{table}

%% file: sections/5_method.tex
\section{\MakeUppercase{Privacy-Preserving Orthogonal Aggregation}}
\label{sec:method}
\subsection{Motivation}

\begin{figure}[h!]
    \centering
    \includegraphics[width=0.5\linewidth]{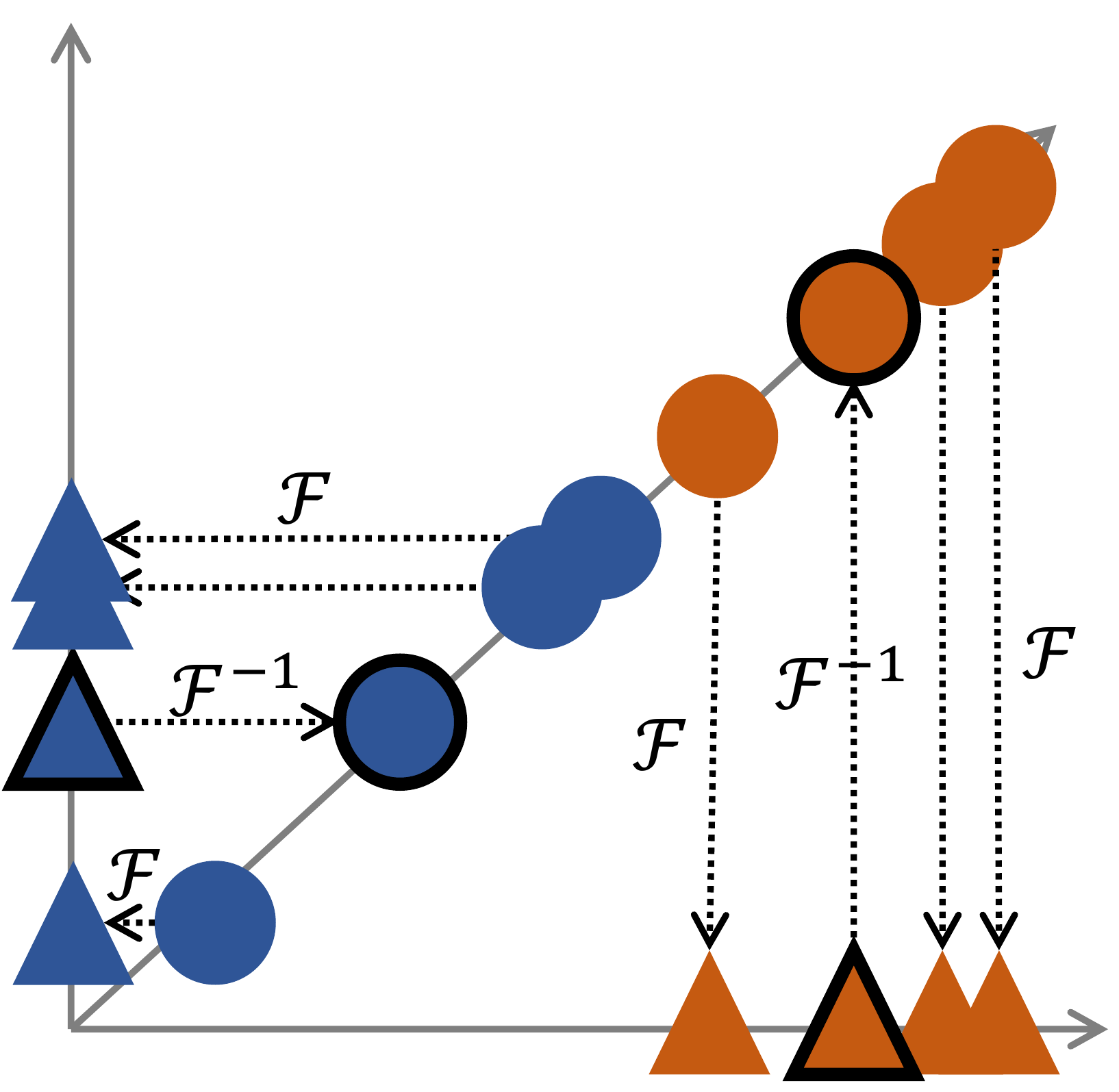}
    \caption{A diagram of OA. Red represents females, and blue represents males. Circle represents the item embedding in low dimensional (original) space, and triangle represents the item embedding in high-dimensional space. Elements with black outlines represent the aggregation results.}
    \label{fig:demo-oa}
\end{figure}

From Eq.~\eqref{eq:f2mf_loss}, it is apparent that while the $\mathcal{L}_{rec}$ does constrain the model to produce similar recommendation performance for two distinct groups, it's \textbf{unfair} to the superior group. 
Therefore, we explore an approach to separately aggregate the model updates for each gender group, \textbf{allowing both male and female to adequately learn their respective features without bias from the other group} (please refer to Sec.~\ref{sec:discussion} for the discussion on the commonalities between groups). However, in the absence of external datasets and relying solely on user history records, it is \textit{impossible to use clustering followed by independent aggregation}. Most crucially, it is imperative that gender attributes are well protected. To tackle this challenge, we propose a solution involving the definition of a bijection function $\mathcal{F}:\mathbb{R}^d\rightarrow \mathbb{R}^{2d}$ and its inverse $\mathcal{F}^{-1}:\mathbb{R}^{2d}\rightarrow \mathbb{R}^{d}$. $\mathcal{F}$ maps the vectors of different genders into two mutually orthogonal subspaces in a higher-dimensional space, as shown in Fig.~\ref{fig:demo-oa}. The server aggregates the vectors in the usual manner, and users can utilize $\mathcal{F}^{-1}$ to obtain the aggregation results corresponding to their own gender.

\subsection{Orthogonal Aggregation}
\label{sec:oa}
Here, we provide a detailed description of the orthogonal aggregation scheme. For the privacy protection strategies and the complete protocol flow of PPOA, please refer to Sec.~\ref{sec:putTogether}.

Let $\alpha$ denote the male and $\beta$ denote the female. The original vector of the $j$-th user in group $i$ is $\bm{\Theta}^{i(j)}$, where $i \in \{ \alpha, \beta \}$, $j \in [1,n_i]$. Randomly select integers $p,q$ within a small range, and define two orthogonal vectors with equal norm as male and female \textit{Attribute Vectors}, respectively:
\begin{equation}
    \bm{\nu}_{\alpha} = (p,q),\bm{\nu}_{\beta} = (-q,p).
\end{equation}
We define $\mathcal{F}$ as follows:
\begin{equation}
        \mathcal{F}\left( {\bm{\nu}_{i},\bm{\Theta}^{i{(j)}}} \right)       
        = \left( \theta_{1}^{i(j)} \bm{\nu}_{i},\ldots, \theta_{d}^{i(j)} \bm{\nu}_{i} \right) = \bm{\vartheta}^{i(j)}.
\end{equation}

After all users employ $\mathcal{F}$ to map the original vectors to their respective subspaces, they transmit $\left(\bm{\vartheta}^{i(j)} || \bm{\nu}_{i}\right)$ to the server. The server obtains the sum of these two parts separately, i.e., $\mathbf{W_{vec}}$ and $\mathbf{W_{num}}$, and sends $\left(\mathbf{W_{vec}}  ||\mathbf{W_{num}}  \right)$ to the users. We define $\mathcal{F}^{- 1}$ as follows:
\begin{equation}
\mathcal{F}^{- 1}\left( {\bm{\nu}_{i},\mathbf{W}} \right)                                                              = \left( \left(\mathbf{W}_1 \cdot \bm{\nu}_{i}\right),\ldots, \left(\mathbf{W}_d \cdot \bm{\nu}_{i} \right)\right),
\end{equation}
where ``$\cdot$'' denotes the \textit{Dot Product}. Users obtain the average aggregation vectors of their respective groups by:
\begin{equation}
    \begin{aligned}
        Num_{i}             & = \frac{1}{p^2+q^2}\mathcal{F}^{- 1}\left( {\bm{\nu}_{i},\mathbf{W_{num}}} \right),\\
        \mathbf{Avg_{vec_i}} & = \frac{1}{\left(p^2+q^2 \right)Num_{i}}\mathcal{F}^{- 1}\left({\bm{\nu}_{i},\mathbf{W_{vec}}}\right).
    \end{aligned}
\end{equation}
For the correctness and the generalization of OA to $v$ types of groups, we have the following theorem:
\begin{theorem}
\label{theorem:agg_vec}
    $\mathcal{U}$ is a user set which can be classified into $v$ groups based on a sensitive attribute, and the number of users of group $i$ is $n_{i}$ $(i\in [1,v])$. The original vector of the $j$-th user in group $i$ is $\bm{\Theta}^{i(j)}$ $(i\in[1,v],j\in[1,n_{i}])$. The attribute vector of group $i$ is $\bm{\nu}^{i} = \left( {\nu_{1}^{i},\nu_{2}^{i},\ldots,\nu_{v}^{i}} \right)$, requiring attribute vectors to be pairwise unequal and for any two attribute vectors $\bm{\nu}_{e},\bm{\nu}_{f}$ satisfy:
    \begin{equation}
        \bm{\nu}^{e} \cdot \bm{\nu}^{f} = {\sum\limits_{i = 1}^{v}{\nu_{i}^{e}\nu_{i}^{f}}} = \left\{ \begin{aligned}
             & 0,e = f        \\
             & \mu,otherwise
        \end{aligned} \right.,
    \end{equation}
    where $\mu$ is a fixed constant.
    The bijective functions are $\mathcal{F}( \cdot )$ and $\mathcal{F}^{- 1}(\cdot)$.
    For any $g\in[1,v]$, satisfy:
    \begin{equation}
        \mathcal{F}^{- 1}\left( {\bm{\nu}^g,{\sum\limits_{i = 1}^{v}{\sum\limits_{j = 1}^{n_{i}}{\mathcal{F}\left( {\bm{\nu}^{i},\bm{\Theta}^{i{(j)}}} \right)}}}} \right) = \mu{\sum\limits_{j = 1}^{n_{g}}\bm{\Theta}^{g{(j)}}}.
    \end{equation}
\end{theorem}
The proof of the theorem is provided in Appendix~\ref{app:theo1}.

\subsection{Put It All Together}
\label{sec:putTogether}
To protect vector privacy, a trusted third party $\mathcal{TTP}$ provides users with the mask vectors required for SecAgg~\cite{Bonawitz2017PracticalSA}, which ensures that the user's data vectors and attribute vectors, once masked, become indistinguishable from random vectors, thereby mitigating the privacy leakage risks mentioned in Sec.~\ref{sec:f2mf_attack}. Moreover, during the aggregation process, the server automatically cancels out all masks, thus, aside from information about the aggregate result, it is unable to access any individual user's private data. In the aggregation process, the communication overhead is reduced using the quantization technique introduced in Sec.~\ref{sec:sapQ}.

Fig.~\ref{protocol:oa} describes the comprehensive procedural flow of PPOA. Upon completion of the protocol, both the male and female groups obtain the updated model outcomes and then perform the next training iteration.

\input{tab/protocol.tex}

\begin{remark}
    \textbf{Privacy.} During the execution of PPOA, the server is unable to access the user's model updates and sensitive attributes.
\end{remark}
The secure aggregation scheme we employ ensures that the vectors received by the server from users are indistinguishable from random vectors.

\begin{remark}
    \textbf{Computation overhead.} The computational overhead of orthogonal aggregation, which solely involves lightweight scalar multiplications and dot products, is negligible.
\end{remark}
The theoretical additional computational complexity of PPOA is $O(d)$. For the empirical analysis of the computational overhead, please refer to Sec.\ref{sec:ablation}.

\begin{remark}
    \textbf{Communication cost.} In theory, PPOA does not increase the communication cost of users.
\end{remark}
We acknowledge that orthogonal aggregation does increase the number of parameters transmitted by users to the server from $d$ to $2d$, doubling the communication overhead. To this end, we employ quantization to eliminate the additional overhead. For example, if $h=16$, the original 32-bit floating-point number for each parameter is converted to a 16-bit integer, reducing communication overhead to the original level. We explore the impact of quantization on the accuracy in Sec.~\ref{sec:ablation}.

\begin{figure*}[t!]
    \centering
    \includegraphics[width=0.85\textwidth]{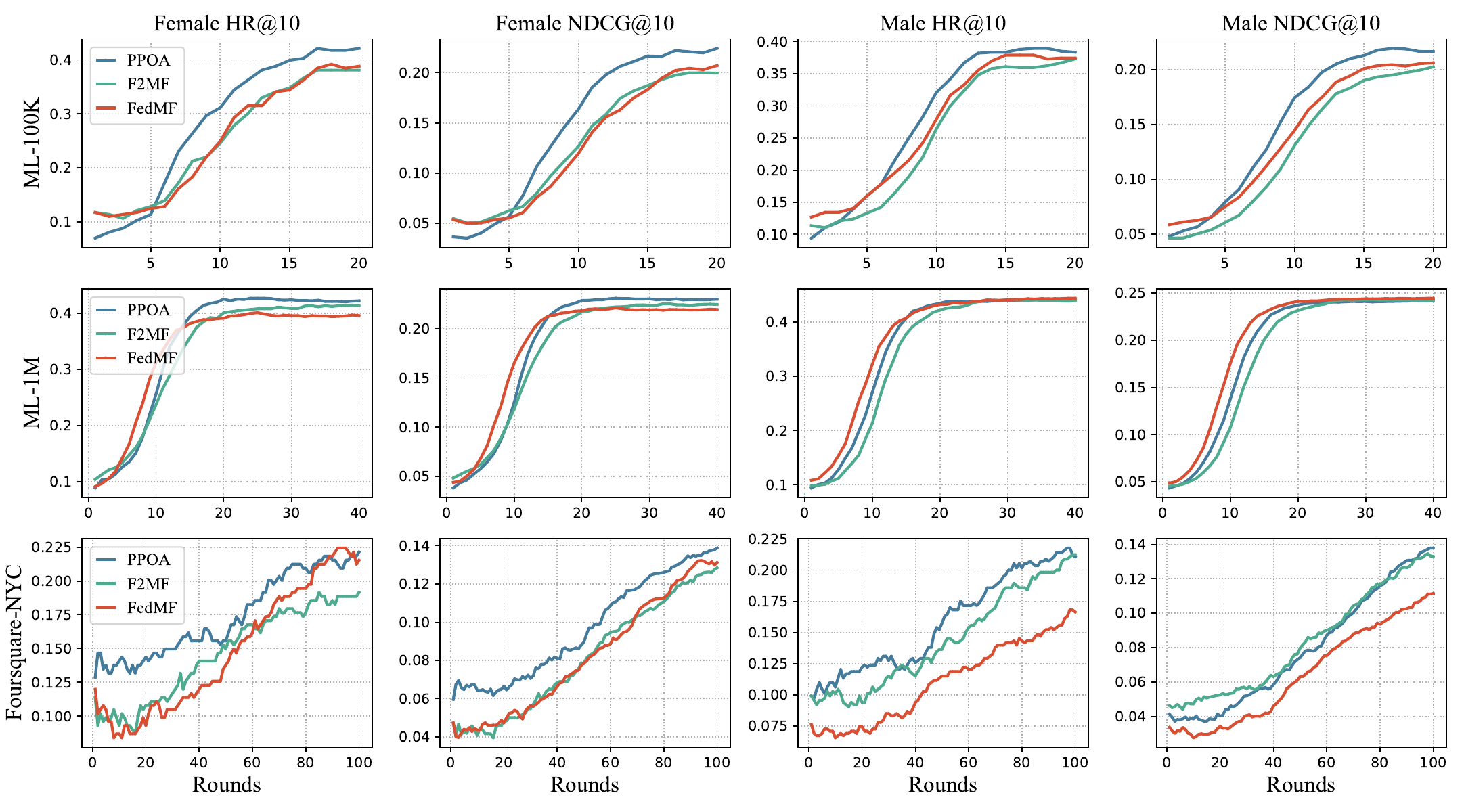}
    \caption{The respective recommendation performance (w.r.t. HR@10 and NDCG@10) for male and female groups compared to the baseline.}
    \label{fig:maincomp}
\end{figure*}

%% file: tab/protocol.tex
\begin{figure}[htbp]
\begin{center}
\begin{tabular}{|p{0.95\linewidth}|}
\hline
\multirow{2}* {\centerline{\textbf{Privacy-Preserving Orthogonal Aggregation Protocol}} }                                                                                                                                                                                           \\
                {} \\
    \textbf{Participants:} Male user set $\mathcal{U}_{\alpha}$ with $n_{\alpha}$ users; Female user set $\mathcal{U}_{\beta}$ with $n_{\beta}$ users; The server $\mathcal{S}$ and $\mathcal{TTP}$.\\
    \textbf{Input:}
    Original vector for each user $\bm{\Theta}^{i(j)}$, $i \in \{ \alpha, \beta \}$, $j \in [1,n_i]$.\\
    \textbf{Output:}
    $\frac{1}{n_{\alpha}}{\sum\limits_{j = 1}^{n_{\alpha}}\bm{\Theta}^{\alpha(j)}}$ and $\frac{1}{n_{\beta}}{\sum\limits_{j = 1}^{n_{\beta}}\bm{\Theta}^{\beta(j)}}$.

    \begin{itemize}[leftmargin=*]
        \item $\mathcal{TTP}:$
              \begin{algorithmic}[1]
                  \State randomly generates $p,q\in \mathbb{Z}^*$, designates $\bm{\nu}_{\alpha}=(p,q)$ as the attribute vector of female and $\bm{\nu}_{\beta}=(-q,p)$ as the attribute vector of male and publicly discloses $\bm{\nu}_{\alpha},\bm{\nu}_{\beta}$.
                  \State generates random mask vectors $\bm{\xi}_{vec}^{i(j)}$ and $\bm{\xi}_{num}^{i(j)}$ for each user based on the secure aggregation scheme and dispatches them to the respective users.
              \end{algorithmic}
        \item Each user $u^{i(j)}$, $i \in \{ \alpha,\beta \}$, $j \in [1,n_i]:$
              \begin{algorithmic}[1]
                  \State computes $\bm{\vartheta} ^{i(j)}_{vec} = \mathcal{F} \left( \bm{\nu}_i, \mathcal{Q}_h\left({\bm{\Theta}^{i(j)}}\right)\right) + \bm{\xi}_{vec}^{i(j)}$ and $\bm{\vartheta} _{num} ^{i(j)} = \bm{\nu}_i + \bm{\xi}_{num}^{i(j)}  $.
                  \State sends $\left( \bm{\vartheta} ^{i(j)}_{vec} || \bm{\vartheta} ^{i(j)}_{num}\right)$ to $\mathcal{S}$.
              \end{algorithmic}
        \item $\mathcal{S}:$
              \begin{algorithmic}[1]
                  \State computes $
                      \mathbf{W}_{vec} = {\sum\limits_{j = 1}^{n_{\alpha}}\bm{\vartheta} _{vec}^{\mathbf{\alpha}{(j)}}} + {\sum\limits_{j = 1}^{n_{\beta}}\bm{\vartheta} _{vec}^{\mathbf{\beta}{(j)}}}$ and $\mathbf{W}_{num} = {\sum\limits_{j = 1}^{n_{\alpha}}\bm{\vartheta} _{num}^{\mathbf{\alpha}{(j)}}} + {\sum\limits_{j = 1}^{n_{\beta}}\bm{\vartheta} _{num}^{\mathbf{\beta}{(j)}}}$.
                  \State sends the calculation results to all users.
              \end{algorithmic}
        \item Each user $u^{i(j)}$, $i \in \{ \alpha,\beta \}$, $j \in [1,n_i]:$
              \begin{algorithmic}[1]
                  \State computes $
                      {Num}_{i} = \frac{1}{p^{2} + q^{2}}\left( {\bm{\nu}_{i} \cdot \mathbf{W}_{num}} \right)$.
                  \State outputs $
                      {\mathbf{A}\mathbf{v}\mathbf{g}}_{vec_i} = \frac{1}{\left( {p^{2} + q^{2}} \right){Num}_{i}} \mathcal{Q}_h^{-1}\left(\mathcal{F}^{- 1}\left( {\bm{\nu}_{i},\mathbf{W}_{\mathbf{vec}}} \right)\right) = \frac{1}{n_{i}}{\sum\limits_{j = 1}^{n_{i}}\bm{\Theta}^{i(j)}}$.
                \end{algorithmic}
    \end{itemize}
    \\\hline
\end{tabular} 
\end{center}
\caption{The workflow of PPOA.}
\label{protocol:oa}
\end{figure}

%% file: sections/6_evalution.tex
\section{\MakeUppercase{Evaluation}}
\label{sec:evaluation}
\subsection{Experimental Settings}
\label{sec:settings}
\textbf{Datasets and Evaluation Metrics. }We utilized three real-world datasets: ML-100K\footnote{https://grouplens.org/datasets/movielens/}, ML-1M~\cite{harper2015movielens},  Foursquare-NYC\footnote{https://sites.google.com/site/yangdingqi/home/foursquare-dataset}~\cite{yang2014modeling}, taking gender as the sensitive attribute. Ratings in these datasets, originally spanning from 1 to 5, are converted to implicit feedback, with all ratings above 0 adjusted to 1. The characteristics of these datasets are detailed in Tab.~\ref{tab:datasets_summary}. Users with fewer than 10 records are excluded, and the datasets are split using a \textit{leave-one-out}~\cite{he2017neural} approach for training and testing. We employ the \textit{Hit Ratio} (HR) and \textit{Normalized Discounted Cumulative Gain} (NDCG) as metrics to assess the performance of the recommendations, with higher values indicating superior effectiveness. It is important to note that the performance for a specific group is evaluated based on the average of all user metrics within that group. When assessing the model's overall performance, we calculate the average of all group performances, rather than an average across all individual users.
\input{tab/dataset}

\textbf{Baseline and Implementation Details. }We compare PPOA with the following federated recommendation models:
\begin{itemize}[leftmargin=*]
    \item \textbf{FedMF}~\cite{chai2020secure}: The implementation of matrix factorization~\cite{koren2009matrix} in the federated setting, which does not take group fairness into consideration.
    \item \textbf{F2MF}~\cite{liu2022fairness}: The state-of-the-art fairness-aware FRS. Fairness constraints is incorporated to ensure that the model's performance remains consistent across different groups.
\end{itemize}
Since no additional data sources are employed, and both F$^2$PGNN and F2MF utilize the same fairness control methods, F$^2$PGNN is not considered as a baseline. In all methods, the length of user and item embeddings is set to 32, the batch size is 256, the learning rate is 0.001, and the number of local training epoches is set to 3. For F2MF, the parameters $\lambda=0.5$ and $\rho=1$ are the optimal values from the original study. In PPOA, the quantization parameter $h$ is set to 16, and the secure aggregation scheme employed is SecAgg~\cite{Bonawitz2017PracticalSA}. We run on a Linux workstation with 32GB of RAM and an Intel (R) Xeon (R) Gold 6246R CPU @ 3.40GHz. We use 1 NVIDIA GeForce RTX 3090 GPU with 24GB RAM only for model training, excluding the aggregation process. Each experiment is repeated 5 times, and the average results are reported.

\subsection{Performance Comparison}
\label{sec:maincom}
\textbf{Recommendation Performance. }
As depicted in Fig.~\ref{fig:maincomp}, it is evident that PPOA outperforms other baselines for both genders, achieving state-of-the-art results. Specifically, in the ML-100K dataset, PPOA performs an NDCG@10 of 0.2243 for females, followed by FedMF at 0.2072, achieving an improvement of 8.25\%. For males, PPOA records an NDCG@10 of 0.2192, with FedMF following at 0.2061, marking a 6.36\% improvement. The overall improvement for all users in ML-100K is 7.30\%. F2MF typically learns more slowly, as seen in cases like males in ML-100K and ML-1M. This is attributed to their methodology to achieve uniform performance across groups, which consequently restricts the learning pace of the dominant group. 

\textbf{Group Fairness.} The interference between groups is more significant for the female. Since PPOA prevents this interference, allowing them, especially the female, to adequately learn their own characteristics, it improves group fairness.
Although PPOA does not perform well in fairness on ML-100K, it exhibits the best recommendation performance compared to baselines on this dataset, regardless of gender, as shown in Fig.~\ref{fig:maincomp}.

\input{tab/fairness_com}

\subsection{Ablation Study}
\label{sec:ablation}

\begin{figure}[h!]
    \centering
    \includegraphics[width=0.85\linewidth]{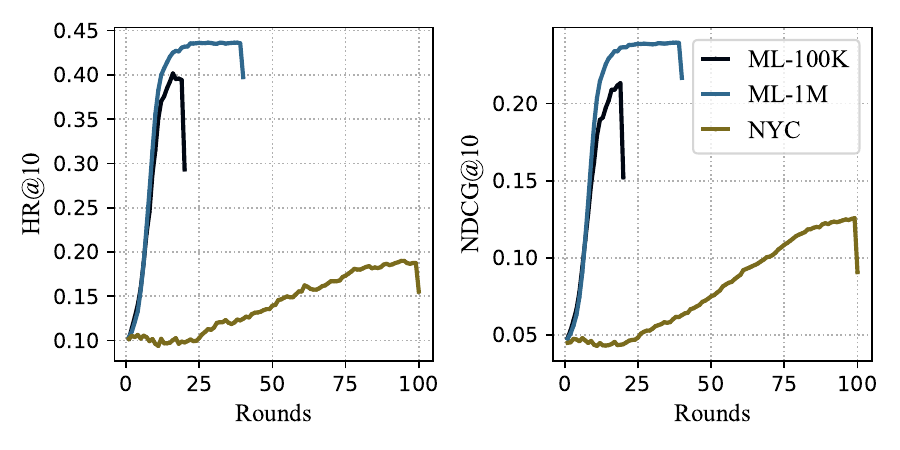}
    \caption{The overall performance of PPOA across three datasets when swapping male and female model parameters in the final evaluation.}
    \label{fig:exchange}
\end{figure}

\textbf{Swap Model Parameters. }As shown in Fig.~\ref{fig:exchange}, when we swap male and female model parameters during the final evaluation, there is a significant decline in user performance, with the NDCG@10 for ML-100K dropping by 28.62\%. This confirms that PPOA helps male and female groups learn parameters that preserve their respective group preferences. Additionally, it reflects the heterogeneity of data between male and female groups.

\begin{figure}[h!]
    \centering
    \includegraphics[width=0.85\linewidth]{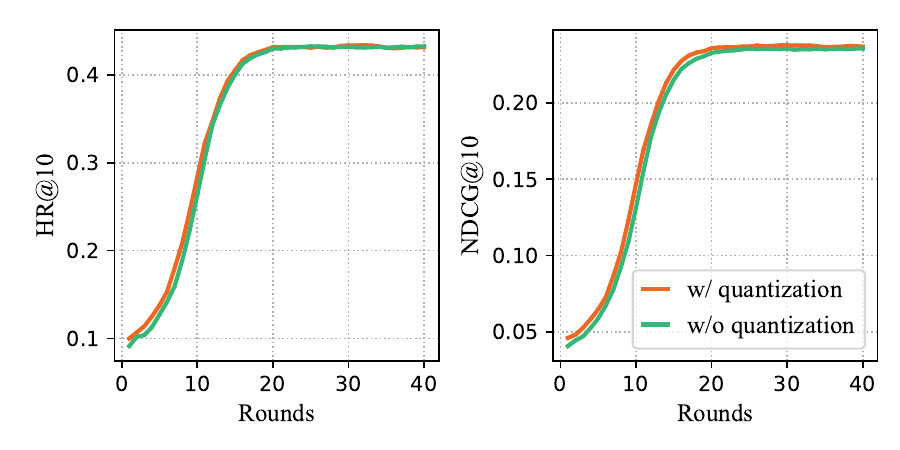}
    \caption{The overall performance of PPOA on ML-1M with and without the quantization.}
    \label{fig:qu-acc}
\end{figure}

\textbf{Impact of Quantization on Accuracy. }As shown in Fig.~\ref{fig:qu-acc}, when $h=16$, the impact of quantization on PPOA accuracy is negligible. Therefore, employing quantization techniques cancels out the communication overhead of PPOA without performance impact.

\begin{figure}[h!]
    \centering
    \includegraphics[width=0.8\linewidth]{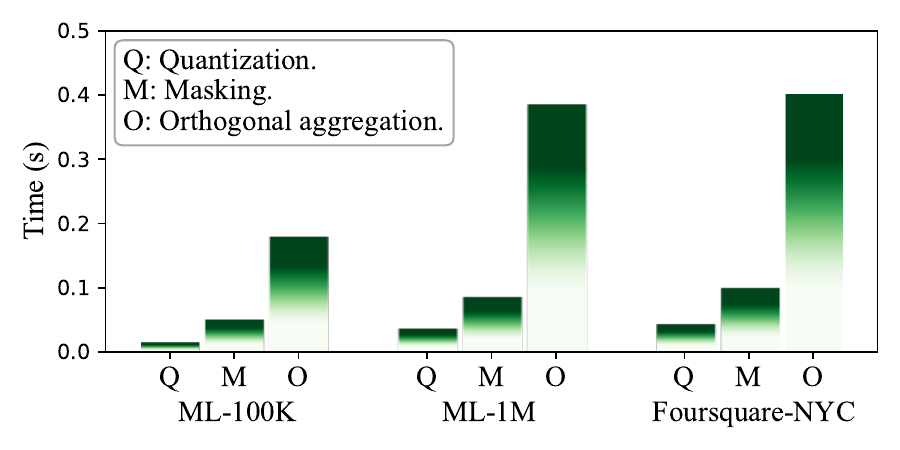}
    \caption{Various computational costs brought by PPOA on three datasets.}
    \label{fig:computation}
\end{figure}

\textbf{Computational Overhead. }Quantization related calculations can be directly operated on the entire vector using $Numpy$~\cite{harris2020array}, resulting in fast computation. The masking operation includes converting floating-point numbers to elements in a finite field and adding masks. In our experiments, the length of the original vector is $32m$, where $m$ is the number of items, so the computational cost for users in the three datasets is different. But even for the dataset with the highest number of items, the additional computational cost that PPOA brings to users does not exceed 1 second.

\begin{figure}[h!]
    \centering
    \includegraphics[width=\linewidth]{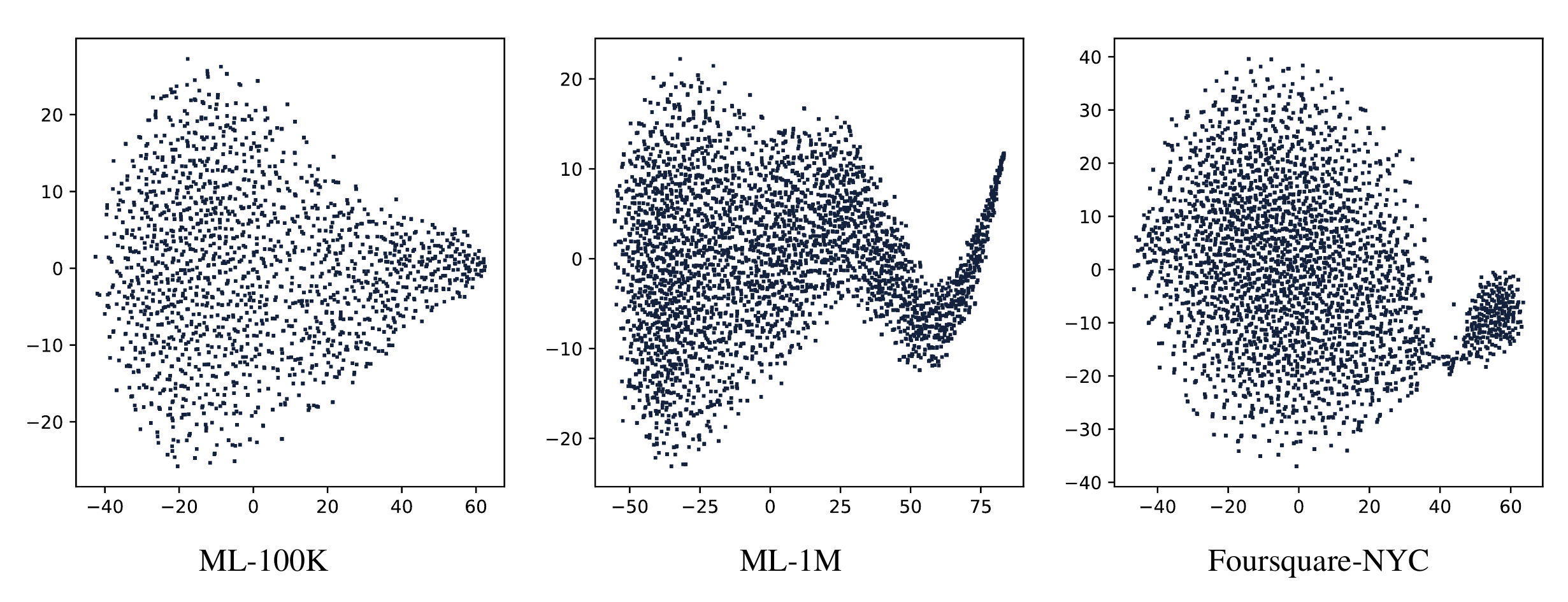}
    \caption{The t-SNE visualization of item embeddings of FedMF on 3 datasets.}
    \label{fig:tsne-no}
\end{figure}

\begin{figure}[h!]
    \centering
    \includegraphics[width=\linewidth]{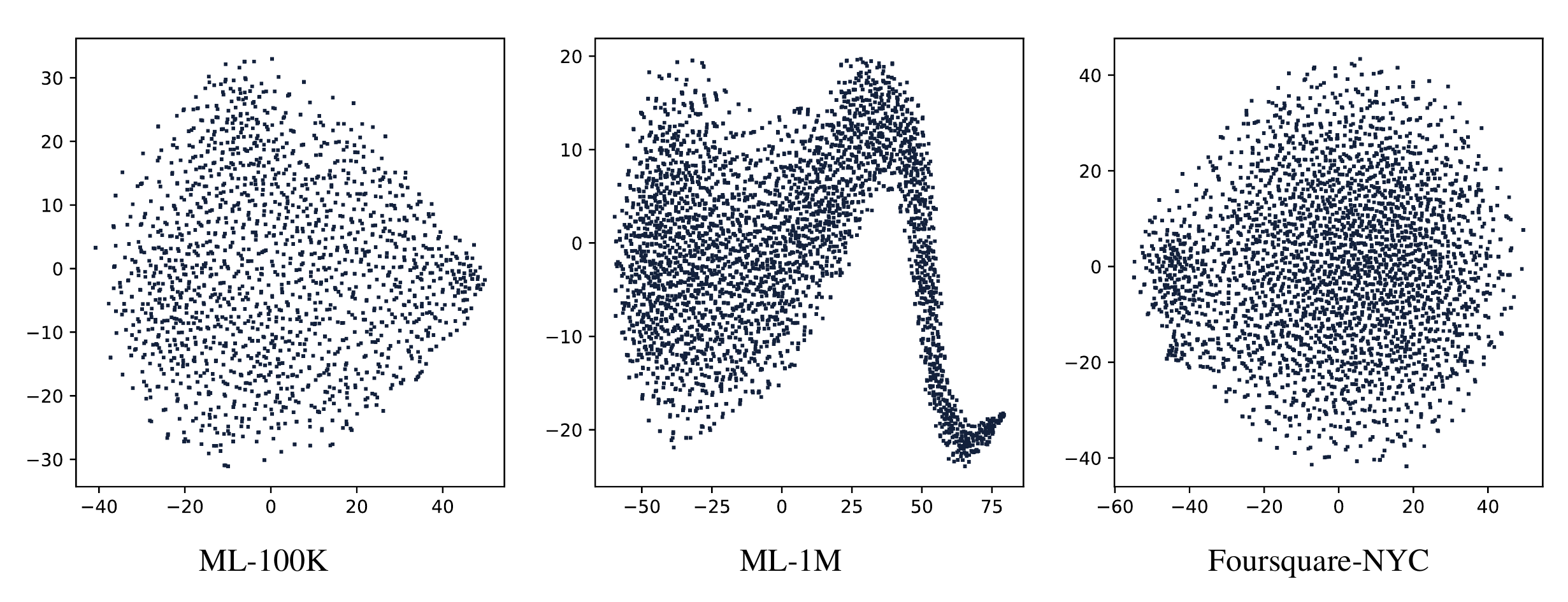}
    \caption{The t-SNE visualization of item embeddings of F2MF on 3 datasets.}
    \label{fig:tsne-fair}    
\end{figure}

\begin{figure}[h!]
    \centering
    \includegraphics[width=\linewidth]{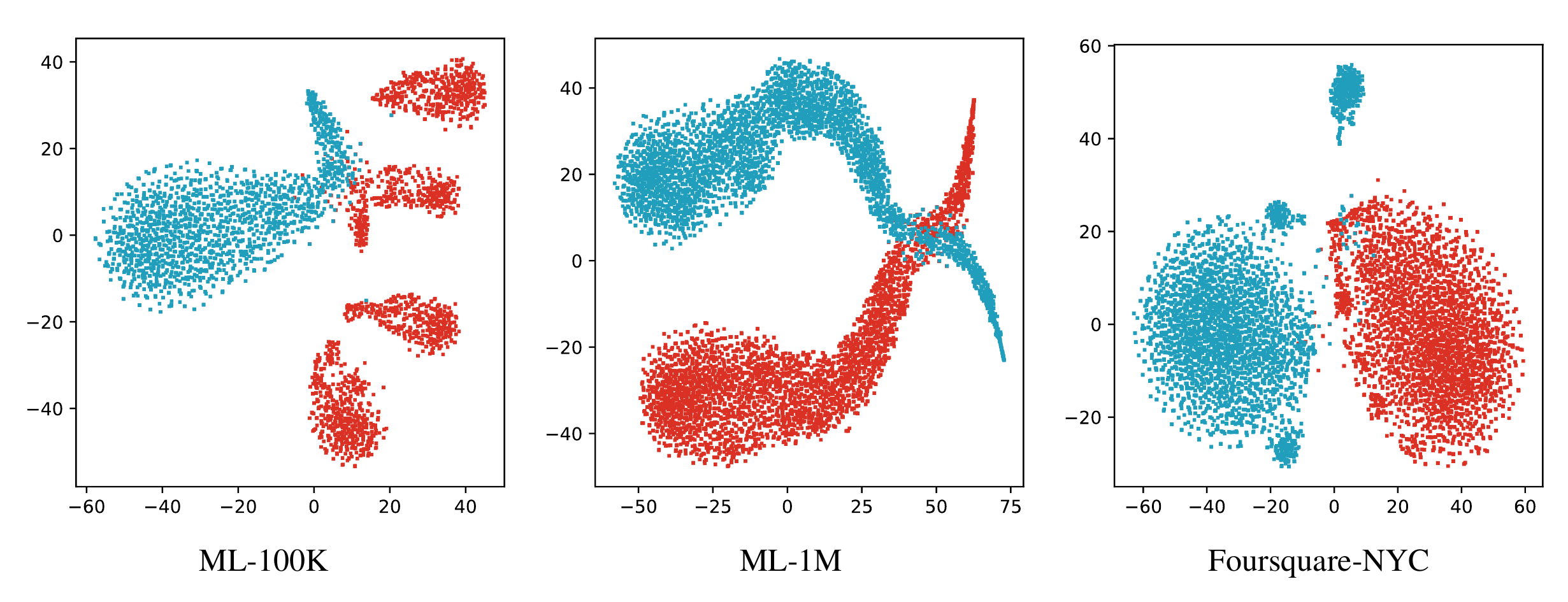}
    \caption{The t-SNE visualization of item embeddings for male and female of PPOA on 3 datasets. Blue indicates items of the male, and red indicates items of the female.}
    \label{fig:tsne}
\end{figure}

\textbf{Visualization. }We use t-SNE~\cite{van2008visualizing} to visualize all item embeddings after the training with the same communication rounds of 3 methods on the same dataset. As shown in Fig.~\ref{fig:tsne-no} and Fig.~\ref{fig:tsne-fair}, the item embeddings learned by male and female users are the same in FedMF and F2MF. As for the performance of PPOA shown in Fig.~\ref{fig:tsne}, the red and blue parts often show \textbf{similarities in shape}, which we believe manifests some common characteristics between men and women. And the difference in their spatial distribution is a manifestation of gender difference, corresponding to the gender \textbf{preference disparity} reflected in Fig.~\ref{fig:ex0-genderbias}(c). This confirms the outstanding performance of PPOA in preserving the preferences of various groups.

%% file: tab/dataset.tex
\begin{table}[!t]
\caption{Summary of Datasets.}
\centering
   \scalebox{0.9}{
\begin{tabular}{lrrrr}
\hline
\toprule
\textbf{Datasets}    & \textbf{\#Ratings} & \textbf{\#Users} & \textbf{\#Items} & \textbf{Sparsity} \\
\midrule
ML-100K    & 100,000   & 943     & 1,682   & 93.70\%  \\
ML-1M       & 1,000,209 & 6,040   & 3,706   & 95.53\%  \\
Foursquare-NYC      & 36,342    & 899   & 3,485  & 98.84\%  \\
\bottomrule
\hline
\end{tabular}
}

\label{tab:datasets_summary}
\end{table}

%% file: tab/fairness_com.tex
\begin{table}[t!]

    \caption{
        Fairness comparison. We use $\left|\mathcal{E}_{\text{male}} - \mathcal{E}_{\text{female}}\right|$ to evaluate fairness, where $\mathcal{E}$ is the recommendation metric. The best and second best results are marked in bold and underline.}
    \centering
    \resizebox{\linewidth}{!}{
        \begin{tabular}{ccccccc}
            \toprule
             \multirow{2}{*}{Method}      &  \multicolumn{2}{c}{ML-100K}     &\multicolumn{2}{c}{ML-1M}      &  \multicolumn{2}{c}{Foursquare-NYC}            \\
            \cmidrule[0.5pt](lr){2-3} \cmidrule[0.5pt](lr){4-5} \cmidrule[0.5pt](lr){6-7}
            {} & HR@10 &NDCG@10 & HR@10 &NDCG@10 & HR@10 &NDCG@10\\
            \midrule[0.5pt]

            FedMF    & \underline{0.0128}  & \textbf{0.0011}   &0.0429           &0.0234 &0.0564&0.0209\\
            F2MF   & \textbf{0.0078}     & \underline{0.0023}  &\underline{0.0264} &\underline{0.0177} &\underline{0.0207}&\underline{0.0060}\\
            PPOA     &0.0316     &0.0051                    & \textbf{0.0165}  & \textbf{0.0110}  & \textbf{0.0038} & \textbf{0.0010}              \\
            \bottomrule
        \end{tabular}}
    \label{tab:fairness_com}
\end{table}

%% file: sections/7_discussion.tex
\section{\MakeUppercase{CONCLUSION and Discussion}}
\label{sec:discussion}
In this work, we take gender fairness as the starting point to explore group fairness in FRSs and find existing works suffering from model performance degradation and privacy leaks of sensitive attributes. To address these issues, we propose a privacy-preserving orthogonal aggregation approach, PPOA, that can preserve the preferences of each group in federated aggregation, avoiding mutual suppression between groups during training. Experimental results indicate that PPOA effectively preserves the preferences of all groups and outperforms existing fairness-aware FRSs in both recommendation performance and fairness. We briefly discuss some potential extensions of PPOA here and leave them for future work.

\textbf{Taking into Account Commonality among Groups. }
Considering the diversity and complexity of the real world, we believe that users who only learn data from their own group may face challenges in some scenarios. For example, utilizing PPOA and insufficient or extremely imbalanced data, models trained by users, especially those belonging to minority groups, are likely to be under-fitting. To solve this problem, based on PPOA, we take the \textit{Group Fusion Coefficient} $\gamma\in(0,1)$ for an example. After two groups $\alpha$ and $\beta$ obtain aggregated results $\bm{\Theta}_{\alpha}^t=\frac{1}{n_{\alpha}}{\sum\limits_{j = 1}^{n_{\alpha}}\bm{\Theta}^{\alpha(j)}}$ and $\bm{\Theta}_{\beta}^t=\frac{1}{n_{\beta}}{\sum\limits_{j = 1}^{n_{\beta}}\bm{\Theta}^{\beta(j)}}$, users update model parameters by:
\begin{equation}
    \label{eq:theta_update}
    \begin{aligned}
        \bm{\Theta}^{t+1}_{\alpha} & = (1-\gamma) \bm{\Theta}^{t}_{\alpha} +\gamma\bm{\Theta}^{t}_{\beta}, \\
        \bm{\Theta}^{t+1}_{\beta}  & = (1-\gamma) \bm{\Theta}^{t}_{\beta} +\gamma\bm{\Theta}^{t}_{\alpha}. \\
    \end{aligned}
\end{equation}

\textbf{Expanding to Other Attributes. }This paper selects representative gender as the sensitive attribute to explore group fairness in FRSs. Obviously, PPOA can be easily extended to other attributes, and correctness is ensured by Theorem~\ref{theorem:agg_vec}. However, PPOA faces the challenge of communication overhead. In the future, we will explore the combination of other technologies such as \textit{Sparsification}~\cite{lu2023top} to further improve communication overhead.

%% file: sections/ethicalconsiderations.tex
\section*{\MakeUppercase{Ethical Considerations}}
PPOA has no potential negative impact on society. Instead, it can help federal recommendation models achieve better group fairness. And it effectively preserves users' sensitive attributes and data privacy.

%% file: sections/appendix.tex
\appendix
\section{\MakeUppercase{A Plain Method}}
A straightforward approach of obtaining the aggregated values of different groups involves male users appending a $d$-dimensional zero vector to their original vector, while female users append their original vector to a $d$-dimensional zero vector. By employing a secure aggregation scheme, gender privacy can be ensured. After the server aggregates high-dimensional vectors, users can extract the aggregated results for male and female groups by selecting the first $d$ dimensions and the last $d$ dimensions, respectively. However, we do not adopt this method due to the potential vulnerability of the masking process to side-channel attacks~\cite{Spreitzer2018Systematic}, which could lead to the leakage of gender information.

\textbf{Side-Channel Attack}. Attackers can gather side-channel leakage information such as time, power consumption, electromagnetic radiation, sound, heat, radio frequency, and fault output when cryptographic algorithms are executed on a target device~\cite{LeCC08}. By analyzing these information and their correlation with the intermediate computations and states occurring during the operation of cryptographic devices, attackers can potentially reconstruct confidential information. In real-world scenarios, side-channel attacks have been capable of stealing user browsing histories~\cite{weinberg2011still}, decrypting email passwords~\cite{genkin2015get}, etc, \textbf{posing a significant threat to vulnerable edge devices in federated settings}~\cite{wang2021federated,rehman2022federated}. 

In secure aggregation, users need to convert floating-point numbers to a finite field and add masks generated by $\mathcal{TTP}$. If the plain method is used, male users will have noticeable differences in CPU usage and power consumption in the first $d$ dimensions compared to the latter $d$ dimensions which are all zeros, and it is the opposite for female users. Given that $d$ is often large (e.g., $d=3706k, k\in \{16,32,64,128\}$ in ML-1M), these noticeable discrepancies in computational consumption between the front and back ends can be sensitively detected. If an attacker obtains side-channel leakage information through other processes on the device or through some monitoring methods, they can deduce the user's gender. Tab.~\ref{tab:side_attack_ops} summarizes the computational operations involved in the masking process for $d$-dimensional zero and non-zero vectors, including pair-wise and non-pair-wise mask-based state-of-the-art secure aggregation schemes, SecAgg~\cite{Bonawitz2017PracticalSA} and EffiAgg~\cite{Liu2022EfficientDA}. Consequently, if the plain method is employed, the privacy of sensitive attributes remains under threat. Next, we introduce the \textit{Orthogonal Aggregation} (OA) to tackle these issues.
\input{tab/sideattack-ops.tex}
\section{\MakeUppercase{Proofs}}
\subsection{Proof of Lemma 1}
\label{app:lemma1}
\begin{proof}
    We have:
    \begin{equation}
        \operatorname{Pr}(\mu-n \sigma \leq X \leq \mu+n \sigma)=\int_{\mu-n \sigma}^{\mu+n \sigma} \frac{1}{\sqrt{2 \pi} \sigma} e^{-\frac{1}{2}\left(\frac{x-\mu}{\sigma}\right)^2} d x.
    \end{equation}
    Let $y=\frac{x-\mu}{\sigma}$, then:
    \begin{equation}
        \operatorname{Pr}(\mu-3 \sigma \leq X \leq \mu+3 \sigma)=\frac{1}{\sqrt{2 \pi}} \int_{-3}^3 e^{-\frac{y^2}{2}} d y \approx 0.9973.
    \end{equation}
\end{proof}

\subsection{Proof of Theorem 1}
\label{app:theo1}
\begin{proof}
    \begin{equation}
        \begin{aligned}
             & \mathcal{F}^{- 1}\left( {\bm{\nu}^g,{\sum\limits_{i = 1}^{v}{\sum\limits_{j = 1}^{n_{i}}{\mathcal{F}\left( {\bm{\nu}^{i},\bm{\Theta}^{i{(j)}}} \right)}}}} \right)                                                                \\
             & = \mathcal{F}^{- 1}( \bm{\nu}^g,\sum\limits_{i = 1}^{v}\sum\limits_{j = 1}^{n_{i}}( ( {\nu_{1}^{i}\theta_{1}^{i{(j)}},\ldots,\nu_{v}^{i}\theta_{1}^{i{(j)}}} ),\ldots,                                                            \\
             & \quad \quad\quad\quad\quad\quad\quad\quad\quad\quad( {\nu_{1}^{i}\theta_{d}^{i{(j)}},\ldots,\nu_{v}^{i}\theta_{d}^{i{(j)}}} ) ) )                                                                                                 \\
             & = \mathcal{F}^{- 1}( \bm{\nu}^g,( ( {{\sum\limits_{i = 1}^{v}{\sum\limits_{j = 1}^{n_{i}}{\nu_{1}^{i}\theta_{1}^{i{(j)}}}}},\ldots,{\sum\limits_{i = 1}^{v}{\sum\limits_{j = 1}^{n_{i}}{\nu_{v}^{i}\theta_{1}^{i{(j)}}}}}} ),     \\
             & \quad \quad\quad \ldots,( {\sum\limits_{i = 1}^{v}{\sum\limits_{j = 1}^{n_{i}}{\nu_{1}^{i}\theta_{d}^{i{(j)}}}}},\ldots,{\sum\limits_{i = 1}^{v}{\sum\limits_{j = 1}^{n_{i}}{\nu_{v}^{i}\theta_{d}^{i{(j)}}}}} ) ) )              \\
             & = ( \nu_{1}^{g}{\sum\limits_{i = 1}^{v}{\sum\limits_{j = 1}^{n_{i}}{\nu_{1}^{i}\theta_{1}^{i{(j)}}}}} + \ldots + \nu_{v}^{g}{\sum\limits_{i = 1}^{v}{\sum\limits_{j = 1}^{n_{i}}{\nu_{v}^{i}\theta_{1}^{i{(j)}}}}},\ldots,        \\
             & \quad \quad\quad \nu_{1}^{g}{\sum\limits_{i = 1}^{v}{\sum\limits_{j = 1}^{n_{i}}{\nu_{1}^{i}\theta_{d}^{i{(j)}}}}} + \ldots + \nu_{v}^{g}{\sum\limits_{i = 1}^{v}{\sum\limits_{j = 1}^{n_{i}}{\nu_{v}^{i}\theta_{d}^{i{(j)}}}}} ) \\
             & = ( {\sum\limits_{i = 1}^{v}{\nu_{1}^{g}\nu_{1}^{i}{\sum\limits_{j = 1}^{n_{i}}\theta_{1}^{i{(j)}}}}} + \ldots + {\sum\limits_{i = 1}^{v}{\nu_{v}^{g}\nu_{v}^{i}{\sum\limits_{j = 1}^{n_{i}}\theta_{1}^{i{(j)}}}}},\ldots,        \\
             & \quad \quad\quad {\sum\limits_{i = 1}^{v}{\nu_{1}^{g}\nu_{1}^{i}{\sum\limits_{j = 1}^{n_{i}}\theta_{d}^{i{(j)}}}}} + \ldots + {\sum\limits_{i = 1}^{v}{\nu_{v}^{g}\nu_{v}^{i}{\sum\limits_{j = 1}^{n_{i}}\theta_{d}^{i{(j)}}}}})  \\
             & = ( {\sum\limits_{i = 1}^{v}{\nu_{i}^{g}\nu_{i}^{1}{\sum\limits_{j = 1}^{n_{1}}\theta_{1}^{1{(j)}}}}} + \ldots + {\sum\limits_{i = 1}^{v}{\nu_{i}^{g}\nu_{i}^{v}{\sum\limits_{j = 1}^{n_{v}}\theta_{1}^{v{(j)}}}}},\ldots,        \\
             & \quad\quad\quad {\sum\limits_{i = 1}^{v}{\nu_{i}^{g}\nu_{i}^{1}{\sum\limits_{j = 1}^{n_{1}}\theta_{d}^{1{(j)}}}}} + \ldots + {\sum\limits_{i = 1}^{v}{\nu_{i}^{g}\nu_{i}^{v}{\sum\limits_{j = 1}^{n_{v}}\theta_{d}^{v{(j)}}}}})   \\
             & = ( 0 + \ldots + {\sum\limits_{i = 1}^{v}{\nu_{i}^{g}\nu_{i}^{g}{\sum\limits_{j = 1}^{n_{g}}\theta_{1}^{g{(j)}}}}} + \ldots + 0,\ldots,                                                                                           \\
             & \quad \quad \quad \quad \quad 0 + \ldots + {\sum\limits_{i = 1}^{v}{\nu_{i}^{g}\nu_{i}^{g}{\sum\limits_{j = 1}^{n_{g}}\theta_{d}^{g{(j)}}}}} + \ldots + 0 )                                                                       \\
             & =( \mu{\sum\limits_{j = 1}^{n_{g}}\theta_{1}^{g{(j)}}},\ldots,\mu{\sum\limits_{j = 1}^{n_{g}}\theta_{d}^{g{(j)}}})                                                                                                              \\
             & = \mu{\sum\limits_{j = 1}^{n_{g}}\bm{\Theta}^{g{(j)}}}.
        \end{aligned}
    \end{equation}
Due to $\mu$ being a public constant, users can obtain aggregated results for each group separately. 
Obviously, the integration of quantization techniques in Fig.~\ref{protocol:oa} does not affect the validity of the above conclusion.
\end{proof}

\section{\MakeUppercase{Commonality among Groups}}

\begin{figure}[h!]
    \centering
    \includegraphics[width=0.85\linewidth]{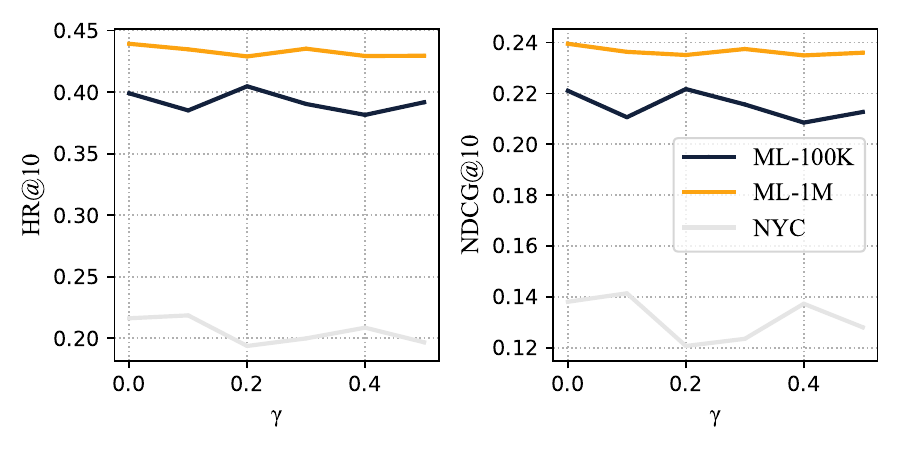}
    \caption{The overall performance of PPOA with various $\gamma$.}
    \label{fig:lam}
\end{figure}

As depicted in Fig.~\ref{fig:lam}, we find that increasing $\gamma$ in the three datasets has limited impact on the performance improvement. This suggests that \textbf{the commonalities between groups play a minimal role in the user learning process}. We surmise that this is because the suppressive effect between groups is more significant than the group commonality effect. In the future work, we will explore the impact of $\gamma$ in the case of less data and extreme imbalance. 

%% file: tab/sideattack-ops.tex
\begin{table}[h!]
    \caption{
        The number of various computational operations required for masking non-zero vectors and zero vectors.}
    \centering
    \resizebox{\linewidth}{!}{
        \begin{tabular}{cccc}
            \toprule
            {Secure aggregation scheme} & \multicolumn{1}{|c} {SecAgg} & \multicolumn{2}{|c}{EffiAgg}                          \\
            \midrule[0.5pt]
            {Operation}                 & \multicolumn{1}{|c|}{Modular addition}                 & \multicolumn{1}{c|}{Modular exponentiation}                                       & Modular multiplication \\
            \midrule[0.5pt]
            \multicolumn{1}{c|}{Non-zero vector}             & $d$                                                          & $d$                                                    & $d$                    \\
            \multicolumn{1}{c|}{Zero vector}                 & $\mathbf{0}$                                                          & $\mathbf{0}$                                                    & $\mathbf{0}$                    \\
            \bottomrule
        \end{tabular}}
    \label{tab:side_attack_ops}
\end{table}